# Improved CNN-Transformer Network for Neutron Image Super-Resolution

Yuheng Chu[1], Hongli Li[1], Tian Zhang[1*]
([1] Northeast Normal University, Changchun 130024, Jilin, China
Corresponding author. E-mail: zhangt100@nenu.edu.cn)

**Abstract**: The spatial resolution of neutron images is of great importance for material defect diagnosis, structural characterization, and nondestructive testing. However, owing to limitations in neutron beam quality, scintillator conversion efficiency and light scattering, detector pixel size, and noise levels, conventional neutron images often suffer from speckle noise, blurred edges, and missing details, making them inadequate for high-precision analysis. To address these issues, an improved super-resolution model combining convolutional neural networks (CNNs) and Transformers is proposed. First, to overcome the difficulty of obtaining high-quality and clear training images for neutron imaging, a pretrained image restoration model is fine-tuned using real neutron images and then employed to enhance existing neutron images, thereby providing high-quality reference images for subsequent super-resolution model training. Second, an improved CRAFT-based super-resolution network is developed by incorporating a hybrid prior-guided channel fusion module (HPCF), a mean–standard deviation dual-statistics squeeze-and-excitation block (MS-SEBlock), and a partial convolution feed-forward network (PCFN), with the aim of improving the reconstruction of important information, such as textures and fine details, in neutron images. Simulation results demonstrate that, compared with bicubic interpolation, SRCNN, SRGAN-VGG, and other methods, the proposed network achieves superior performance in both full-reference metrics, including PSNR and SSIM, and no-reference metrics, including AGM and SMD. The visual quality and structural fidelity of the reconstructed images are also significantly improved. This study provides a new approach to improving neutron image quality.

## 1. Introduction

Neutron imaging has significant applications in the field of nondestructive testing (NDT) owing to the high penetrating capability of neutrons and their sensitivity to light elements. In recent years, image super-resolution techniques have been increasingly applied to tasks such as medical diagnosis and industrial inspection [1–3]. Improving spatial resolution is therefore of great importance for extending the capabilities of neutron imaging in nondestructive testing. The spatial resolution of neutron images is not determined by a single factor; rather, it is jointly affected by key components, including neutron source characteristics, detector performance, and collimator design, as well as physical factors such as neutron scattering, particle cross-talk, statistical fluctuations in neutron fluence rate, and gamma-ray contamination [4–6]. Image blurring and noise degradation caused by these factors adversely affect subsequent nondestructive testing and elemental analysis.

Overcoming the resolution limitations at the hardware level requires upgrading large-scale neutron source facilities to increase neutron yield and improve beam quality, as well as developing more advanced scintillator screens and detectors. Although these improvements can significantly enhance imaging quality, they involve substantial costs and are difficult to implement on a wide scale [7–9]. Deep learning has been successfully applied to medical image enhancement, including low-dose CT denoising and MR image restoration [10–13], motivating its application to neutron image super-resolution.

Traditional super-resolution methods mainly include interpolation-based, reconstruction-based, and learning-based approaches [14,15]. However, these methods have certain limitations in recovering details from complex images. In contrast, deep learning can learn image features in a data-driven manner and has demonstrated promising reconstruction capabilities in various fields, including medical imaging [16], defect detection [17], and industrial computed tomography (CT) [18]. It has also opened up new research directions for neutron image super-resolution. Owing to their high reconstruction accuracy and good scalability, deep learning-based methods have been widely applied to super-resolution reconstruction and can be readily integrated with auxiliary tasks such as detection and recognition. Furthermore, these methods can substantially enhance the information content of neutron images [19].

In recent years, deep learning has gradually been introduced into neutron image super-resolution tasks. In 2023, the Bhabha Atomic Research Centre combined SRCNN with the POCS algorithm to achieve super-resolution reconstruction of simulated neutron images, providing an initial demonstration of the feasibility of convolutional neural networks in this field. However, the feature representation capability of the network was limited, and its ability to recover complex textures and high-frequency details remained insufficient [20]. Subsequently, Ma et al. incorporated a VGG-based perceptual loss into SRGAN and evaluated its reconstruction performance using real neutron images. Nevertheless, the adversarial generation process may introduce artificial textures and local artifacts, potentially affecting the fidelity of structural information [21]. In the same year, Ma et al. further proposed a hybrid attention network, RHAN, which enhances high-frequency information through multi-scale features and attention mechanisms. However, its adaptability to the complex degradation characteristics of real neutron images still requires improvement [22]. In 2025, Liu et al. proposed a neutron image super-resolution network based on parallel Transformers and achieved promising reconstruction results on both benchmark datasets and real images. However, its training data consisted primarily of grayscale natural images, resulting in a certain domain gap from real neutron images [23]. In 2026, Wu et al. applied Real-ESRGAN to neutron image reconstruction, improving visual quality and detail representation. Nevertheless, its training data were mainly derived from X-ray images, which differ from neutron images in terms of imaging physics and noise characteristics, potentially limiting the model's generalization capability [24].

In summary, existing studies have made progress in neutron image super-resolution. However, several challenges remain, including insufficient adaptation to real neutron images, the risk of introducing artificial textures, and inadequate recovery of weak edges and high-frequency details. To address these issues, this study proposes an improved Cross-Refinement Attention Fusion Transformer (CRAFT) network tailored to the imaging characteristics of real neutron images. The proposed network aims to enhance local high-frequency detail extraction and global feature modeling, thereby improving detail recovery and structural representation in neutron images.

The main contributions of this study are summarized as follows:

1) An adaptive image restoration-based data preprocessing method is proposed for neutron images. A pretrained image restoration model is fine-tuned using neutron image data to improve the quality of the original images, thereby constructing corresponding reference image data for super-resolution model training.

2) An improved CNN-Transformer hybrid super-resolution reconstruction network is developed based on the CRAFT architecture. To address the insufficient representation of local details and inadequate feature fusion in neutron images, the feature processing structures within the network are enhanced to improve local spatial feature modeling, information fusion between different feature branches, and channel-wise feature representation. The resulting network enables ×4 super-resolution reconstruction of neutron images.

3) Quantitative evaluation and visual comparison experiments are conducted on a self-constructed neutron image dataset, providing preliminary validation of the proposed method's effectiveness in recovering image details and improving neutron image quality.

## 2. Methodology

### 2.1 Neutron Image Restoration and Reference Image Construction

Due to limitations in neutron source intensity, detector performance, and imaging conditions, raw neutron images often suffer from noise, blurring, low contrast, and missing fine details. To construct training data suitable for super-resolution reconstruction, a pretrained image restoration model is first employed to preprocess the raw neutron images, thereby improving image quality and providing reference images for subsequent dataset construction.

In this study, the Adaptive Sparse Transformer (AST-v2) image restoration network is adopted as the initial image restoration model. As illustrated in Fig. 1, the model employs a Transformer-based feature modeling mechanism to extract relevant image information through adaptive sparse self-attention and a feature refinement feed-forward network. Considering the differences in imaging characteristics between neutron images and natural images, the pretrained model is further fine-tuned using neutron image data to improve its adaptability to the noise distributions, edge structures, and texture characteristics of neutron images.

During model fine-tuning, the raw neutron images are first subjected to different types and levels of degradation, including noise interference, blurring, local brightness variations, and contrast variations, to generate the input data required for model training. The network is then initialized using the pretrained weights and fine-tuned specifically on neutron image data. Finally, the images generated by the restoration model are used as high-quality reference images in the super-resolution dataset.

It should be noted that, owing to the lack of physically measured high-resolution reference images strictly corresponding to the raw neutron images, the reference images used in this study are primarily generated by the image restoration model. Therefore, these images should be regarded as algorithmically enhanced high-quality reference data rather than ideal high-resolution images obtained through direct physical measurements.

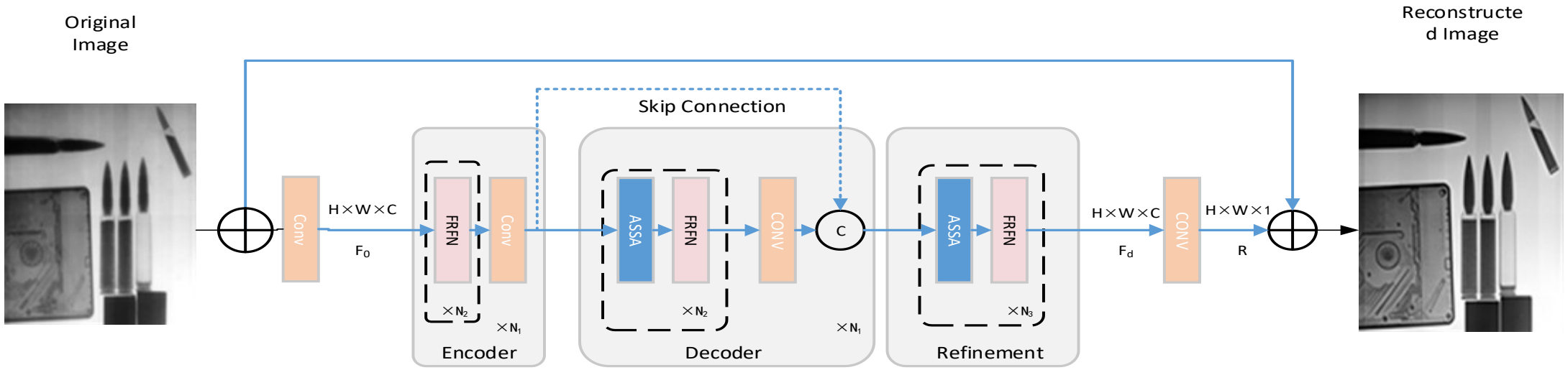

Figure 1: Main Network of AST-v2 Large Model

### 2.2 Dataset Construction

In this study, a neutron image dataset was constructed for super-resolution reconstruction. The images were collected from various representative neutron imaging objects, including aero-engine blades, nuclear fuel assemblies, space reactor heat pipes, special nuclear materials, rocks, batteries, and other mechanical structures. These images contain diverse material boundaries, internal structures, and fine details, making them suitable for evaluating the reconstruction capabilities of super-resolution models across different types of neutron images.

To increase the diversity of the training data, data augmentation was applied to the reference images. The augmentation operations mainly included horizontal flipping, vertical flipping, 90° rotation, grayscale enhancement, Gaussian blurring, brightness adjustment, and contrast adjustment. These operations increase the diversity of training samples and improve the model's adaptability to different image structures and imaging conditions.

Approximately 200 original neutron images were used to construct the dataset. Before patch extraction and data augmentation, the original images were divided into training, validation, and test sets at a ratio of 80%, 10%, and 10%, respectively. Images from the same specimen or acquisition sequence, where applicable, were assigned to the same subset to prevent information leakage. Consequently, no patches derived from the same original image were shared across different subsets.

The image restoration model was fine-tuned exclusively on the training subset. Neither validation nor test images were used for fine-tuning. The resulting model was then frozen and applied independently to each subset to generate algorithmically enhanced reference images.

Within each subset, the reference images were cropped into patches of 256 × 256 pixels, and the corresponding low-resolution patches of 64 × 64 pixels were generated by ×4 downsampling. Data augmentation was applied only to the training subset, while the validation and test subsets were processed without augmentation. Patch extraction and training-set augmentation yielded approximately 10,000 image patches in total. The validation subset was used for model selection, and the test subset was reserved for the final performance evaluation.

### 2.3 CRAFT-Based Super-Resolution Network

In this study, a neutron image super-resolution model is developed based on the CRAFT network. Through convolutional feature extraction, residual feature interaction, and multi-stage feature fusion, the network maps low-resolution images to high-resolution images. Its main components include shallow feature extraction, deep feature extraction, feature fusion, and image reconstruction, ultimately producing ×4 super-resolution results through upsampling. The overall architecture is shown in Fig. 2.

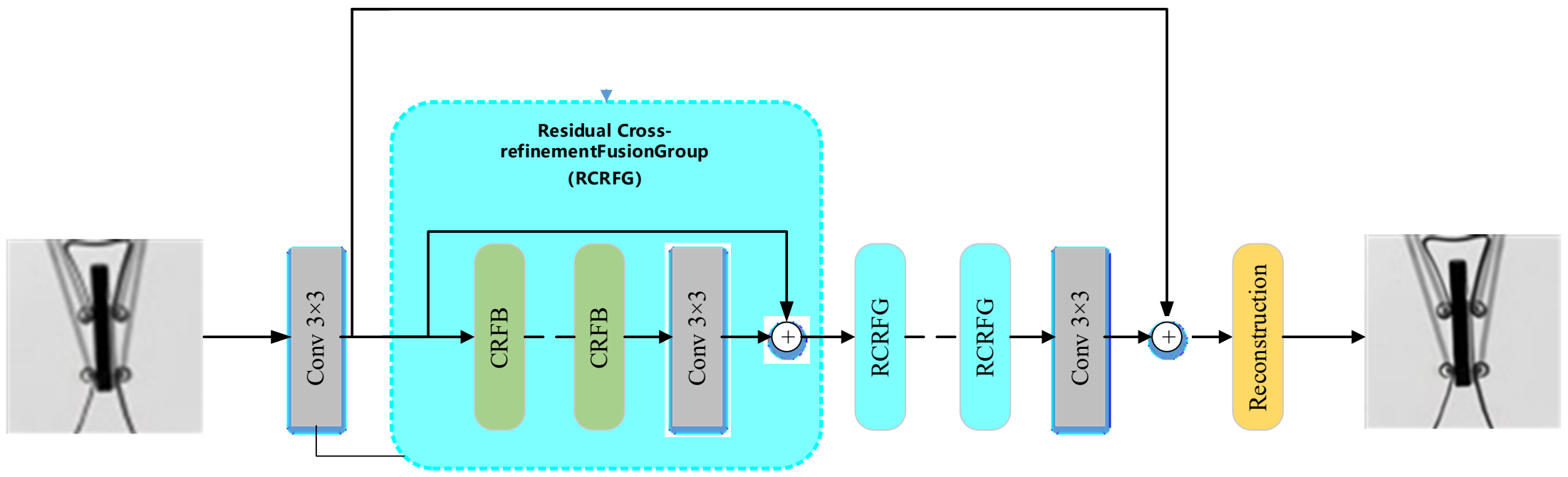

Fig. 2. CRAFT Backbone Network

During neutron imaging, factors such as detector performance, imaging noise, and scattering often lead to low contrast, weak edge responses, and missing details. Consequently, the original network may be insufficient for recovering local details and textures. To address these challenges, this study leverages the specific imaging characteristics of neutron radiography to refine selected feature processing modules within the CRAFT network, thereby enhancing its representational capacity for local details and multi-level features.

Specifically, as shown in Fig. 3, a high-frequency enhancement residual block (HFERB) is employed within the CRFB to extract local high-frequency features, while a shifted rectangular window attention block (SRWAB) is used to model global contextual information. A partial convolution feed-forward network (PCFN) is introduced into the SRWAB to replace the original feed-forward network, enhancing local spatial feature transformation. Furthermore, a hybrid prior-guided channel fusion module (HPCF) is designed to facilitate interactive feature fusion between the HFERB and SRWAB branches. A mean–standard deviation dual-statistics squeeze-and-excitation block (MS-SEBlock) is also incorporated to adaptively recalibrate feature channels within the CRFB, enhancing informative responses and suppressing redundant information.

Through these designs, the network captures both local details and global structural information, providing multi-level feature representations for neutron image super-resolution. Compared with the baseline network, the improved model places greater emphasis on weak edges, fine structures, and high-frequency textures in neutron images.

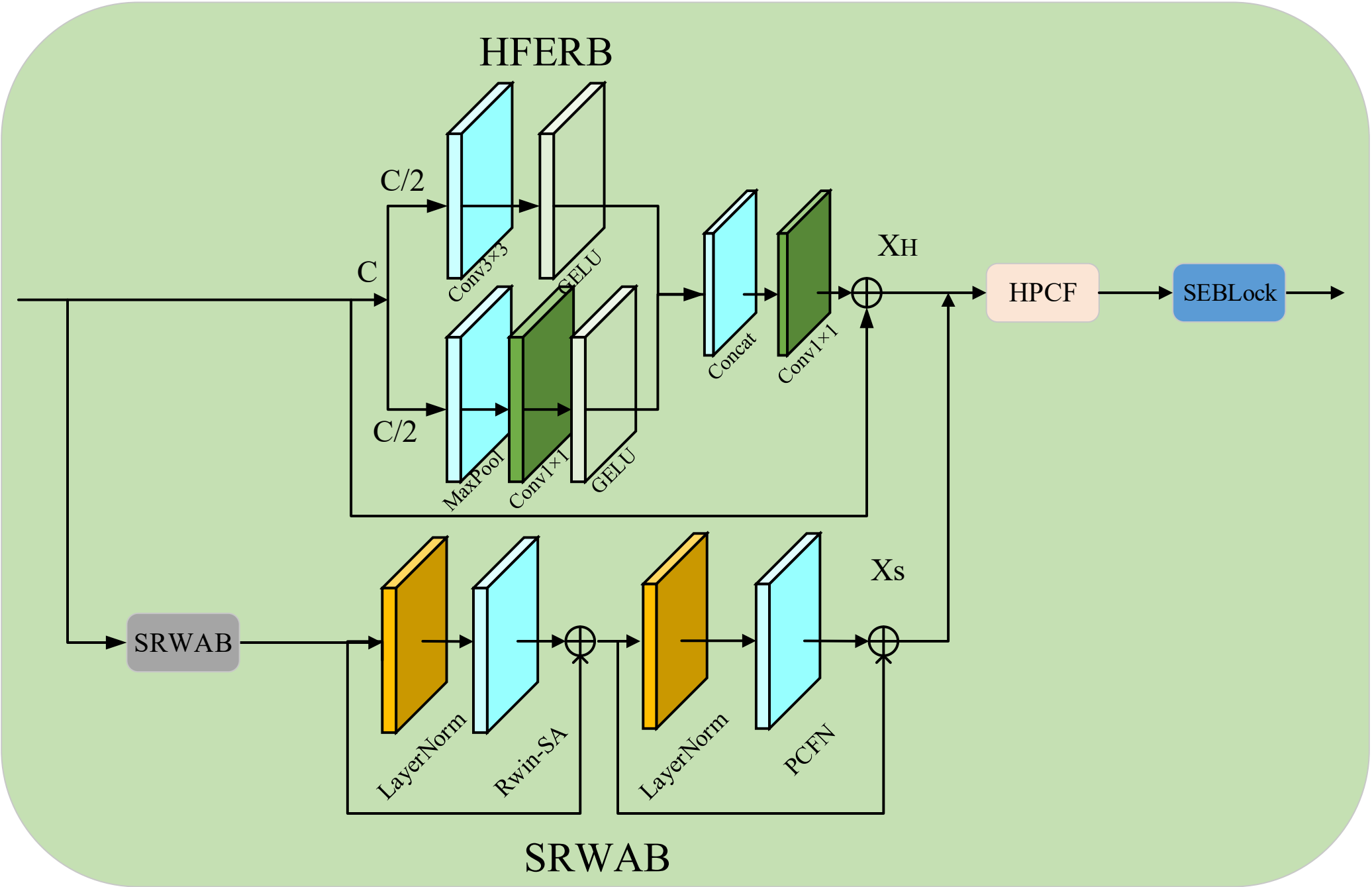

Fig. 3. Internal Architecture of the CRFB

**2.4 Training Objective Function**

To improve the pixel fidelity and edge detail of reconstructed images, a training objective combining the Charbonnier loss and gradient loss is adopted. The Charbonnier loss constrains the pixel-wise differences between the reconstructed and reference images, while the gradient loss constrains local gradient information to enhance the recovery of edges and fine structures. The overall loss function is defined as follows:

$$\mathcal{L}_{\text{total}} = \mathcal{L}_{\text{char}} + \lambda \mathcal{L}_{\text{grad}} \tag{1}$$

where $\mathcal{L}_{\text{char}}$ denotes the Charbonnier loss, $\mathcal{L}_{\text{grad}}$ denotes the gradient loss, and $\lambda$ is the weighting coefficient used to balance the two terms. In this study, $\lambda$ is set to 0.05 to balance pixel reconstruction accuracy and edge detail recovery.

## 3. Experimental Setup and Evaluation Metrics

### 3.1 Experimental Setup

The proposed super-resolution reconstruction network is implemented using the PyTorch deep learning framework and trained on an NVIDIA RTX 4090 GPU. The network parameters are optimized using the Adan optimizer, with an initial learning rate of $2\times10^{-4}$, a batch size of 32, and 500,000 training iterations. The learning rate is adjusted using the MultiStepLR strategy, with each adjustment reducing it to 50% of its previous value.

For image restoration preprocessing, a pretrained model is fine-tuned on neutron images. The AdamW optimizer is used during this stage, with the learning rate adjusted using a cosine annealing schedule. The model is trained for 100 epochs, with the learning rate gradually decreasing from$1\times10^{-5}$ to $1\times10^{-6}$.

Table 1. Fine-tuning training parameters of AST-v2

| Parameter | Setting |
|---|---|
| Input | Degraded neutron images |
| Supervision target | Original neutron images |
| Output | Restored neutron images |
| Loss function | Charbonnier Loss |
| Optimizer | AdamW |
| Learning rate | $1\times10^{-5}$-$1\times10^{-6}$ |
| Learning rate schedule | Cosine annealing decay |
| Data augmentation | Random cropping |
| Training epochs | 100 epoch |

### 3.2 Evaluation Metrics

To quantitatively evaluate the super-resolution results, peak signal-to-noise ratio (PSNR), structural similarity index measure (SSIM), and learned perceptual image patch similarity (LPIPS) are adopted as the primary evaluation metrics. PSNR measures the pixel-wise error between reconstructed and reference images, SSIM evaluates their similarity in terms of luminance, contrast, and structure, while LPIPS quantifies perceptual differences based on distances in deep feature space. In general, higher PSNR and SSIM values and lower LPIPS values indicate greater similarity between the reconstructed and reference images.

In addition, average gradient magnitude (AGM) and standard deviation measure (SMD) are used to assess image details and sharpness. AGM reflects the average intensity of grayscale variations, while SMD describes the dispersion of the grayscale distribution. These metrics provide complementary assessments of image quality from different perspectives. However, their values may also be affected by noise and excessive sharpening; therefore, they should not be used alone as evidence of improvements in actual spatial resolution.

## 4. Experimental Results and Analysis

### 4.1 Quantitative Comparison

The original real neutron radiographic images of turbine blades, which contain noise and other degradations, were first denoised using a Swin Transformer. The denoised images were then processed using the proposed method and other competing methods for comparison. The visual comparison results are shown below.

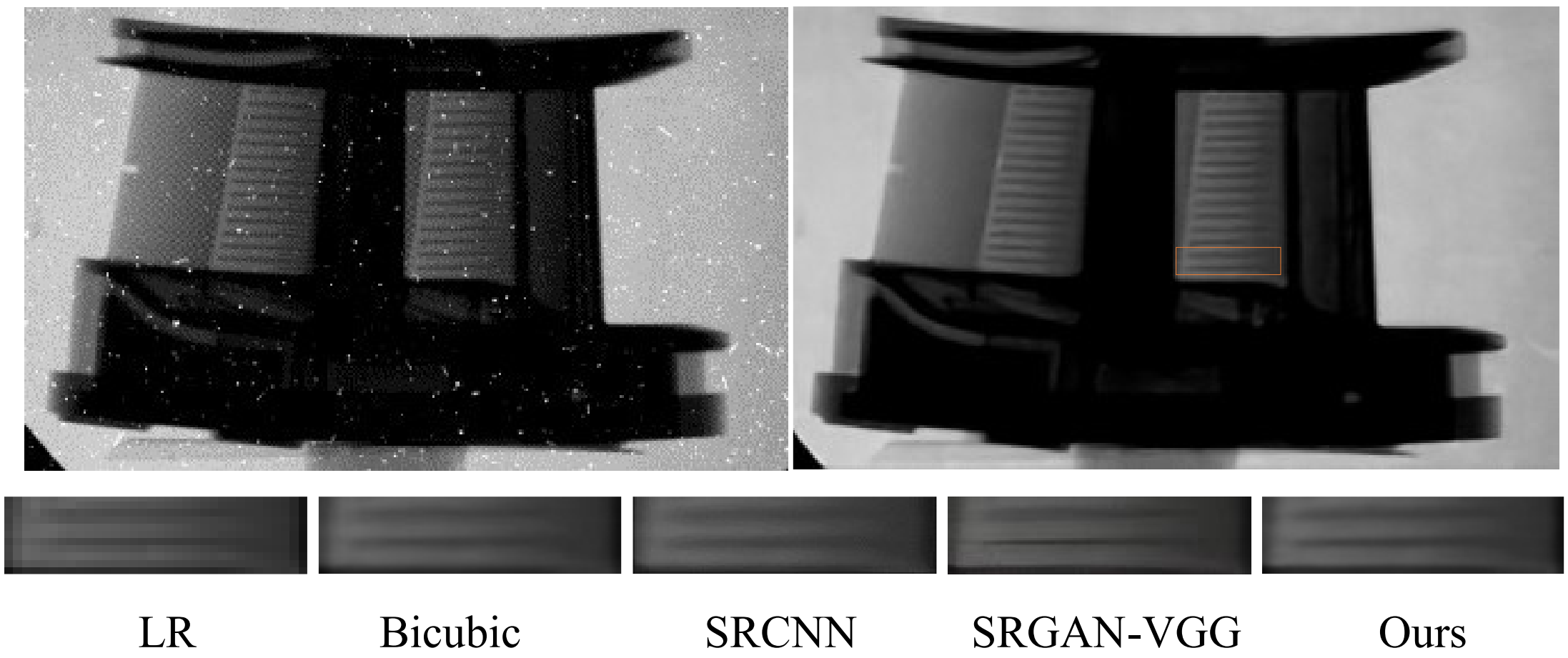

Figure 4. Comparison of ×4 super-resolution reconstruction results for real neutron radiographic images: (a) Neutron radiography (NR) image of a turbine blade acquired at the NEUTRA neutron imaging facility at PSI[25]; (b) real neutron radiographic image after denoising using a Swin Transformer[4].

A comparison of Bicubic, SRCNN, SRGAN-VGG, and Ours shows that the Bicubic results are generally smooth, with blurred horizontal stripe boundaries and limited detail recovery. Compared with Bicubic, SRCNN provides some improvement; however, over-smoothing remains evident, and the separation of the stripe structures is not substantially enhanced. SRGAN-VGG significantly sharpens the stripe boundaries and improves the perceived image sharpness, but it also carries the risk of over-sharpening and introducing artificial textures.

In contrast, Ours achieves a more balanced visual result. The horizontal stripe structures are clearer than those produced by Bicubic and SRCNN, without the excessive sharpening and noise amplification that may occur with SRGAN-VGG. Ours better preserves the grayscale transitions and structural continuity of the original neutron images, achieving a more favorable balance between image sharpness and fidelity. Therefore, Ours delivers superior overall visual quality. In terms of quantitative results, Ours also achieves improvements over the other three methods across all evaluated metrics.

Table 2 Evaluation of Image Quality of Neutron Imaging in Turbine Blades

| Evaluation Metrics / Models | Bicubic | SRCNN | SRGAN-VGG | Ours |
|---|---|---|---|---|
| PSNR | 21.96 | 32.06 | 31.35 | 32.99 |
| SSIM | 0.6568 | 0.9270 | 0.9252 | 0.9368 |
| LPIPS | 0.0352 | 0.0235 | 0.0611 | 0.0184 |
| AGM | 0.0545 | 0.0610 | 0.0580 | 0.0619 |
| SMD | 0.0057 | 0.0062 | 0.0064 | 0.0065 |

For real neutron camera images, the proposed method was further compared with several advanced super-resolution models, including the natural-image super-resolution models SAFMN and WaveMixSR, as well as SRNCT[23], which is specifically designed for neutron image super-resolution reconstruction. As shown in Fig. 5, enlarged local regions of the neutron camera images are used to examine the ability of different methods to recover edge contours, internal textures, and fine structural details. Compared with the other methods, the proposed method restores edge details and internal textures more clearly, providing better discernibility of fine structures and demonstrating stronger detail reconstruction capability.

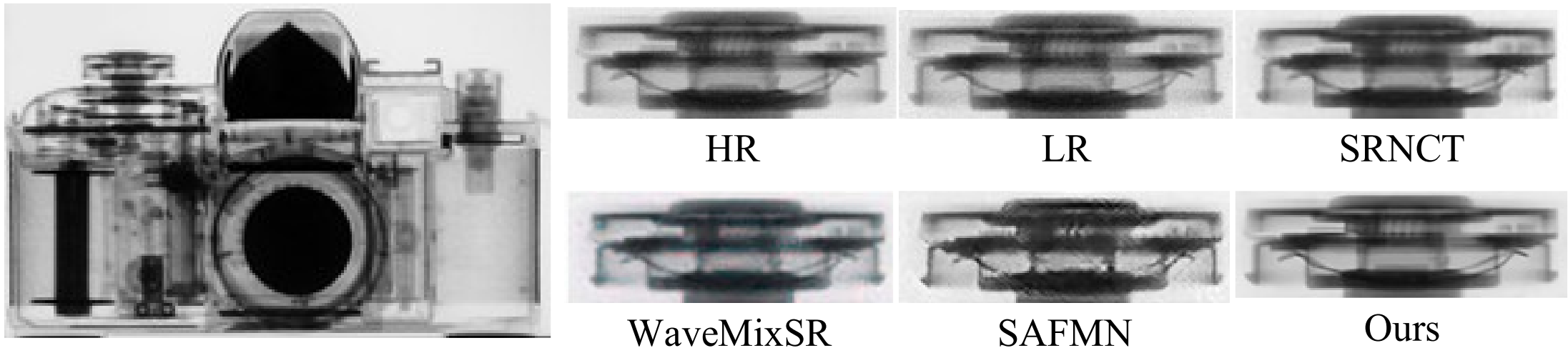

Figure 5: Super-resolution visual effects of neutron cameras compared to other advanced natural super-resolution models

To further evaluate the reconstruction performance of different methods on real neutron camera images, five metrics—PSNR, SSIM, AGM, SMD, and LPIPS—were adopted for quantitative comparison. The results are presented in Table 3. The proposed method achieves a PSNR of 28.06 dB and an SSIM of 0.9336, both higher than those of the competing methods, demonstrating its advantages in pixel-level reconstruction accuracy and structural similarity. Meanwhile, the proposed method obtains the highest AGM (0.2816) and SMD (0.0372), indicating stronger gradient responses and improved detail clarity in the reconstructed images. In terms of perceptual quality, the proposed method achieves an LPIPS of 0.1573, lower than those of the other competing models, suggesting greater perceptual similarity in the feature space. Taken together, the qualitative and quantitative results demonstrate that the proposed network provides effective detail recovery and structural preservation on real neutron camera images, highlighting its potential for neutron image super-resolution tasks.

Table 3. Quantitative Evaluation Results of Different Super-Resolution Methods on Neutron Camera Images

| Evaluation Metrics / Models | SAFMN | WaveMixSR | SRNCT | Ours |
|---|---|---|---|---|
| PSNR | 18.02 | 18.17 | 21.31 | 28.06 |
| SSIM | 0.5879 | 0.6154 | 0.7701 | 0.9336 |
| AGM | 0.2653 | 0.2463 | 0.2465 | 0.2816 |
| SMD | 0.0345 | 0.0323 | 0.0324 | 0.0372 |
| LPIPS | 0.2636 | 0.2448 | 0.2669 | 0.1573 |

**4.2 Discussion and Limitations**

Although the proposed method demonstrates certain advantages in quantitative evaluation and visual comparisons, several limitations remain. First, the high-quality reference images used for training are primarily generated by an image restoration model, and their detail information may be influenced by the preprocessing model. Second, neutron imaging objects and imaging conditions are diverse, while the current dataset remains limited in size and sample distribution. Therefore, the generalization capability of the model across different imaging devices and experimental conditions requires further investigation.

Furthermore, metrics such as PSNR, SSIM, and LPIPS mainly evaluate the similarity between reconstructed and reference images and cannot fully reflect changes in the actual spatial resolution of neutron imaging systems. Therefore, future studies should incorporate more reliable physical reference data, edge response analysis, and modulation transfer function (MTF) measurements to further evaluate the actual resolving capability of the reconstructed images.

## 5. Conclusion

This study investigates deep learning-based super-resolution reconstruction for neutron images, addressing common issues such as noise, blurring, low contrast, and missing details. By using an image restoration model to construct high-quality reference data and introducing targeted feature extraction and fusion structures into the CRAFT architecture, a ×4 super-resolution reconstruction method for neutron images is developed.

Experimental results demonstrate that the proposed method achieves certain advantages in quantitative evaluation and visual comparisons, improving edge clarity and detail representation in neutron images. This study provides an initial approach to applying deep learning-based super-resolution techniques to neutron imaging. Future work will focus on expanding the dataset, improving the physical resolution evaluation framework, and conducting more in-depth investigations into the internal feature interaction mechanisms and generalization capability of the network.